\documentclass[10pt,twocolumn]{article}

\usepackage{cvpr}
\usepackage{times}
\usepackage{epsfig}
\usepackage{graphicx}
\usepackage{amsmath}
\usepackage{amssymb}
\usepackage{graphicx}
\usepackage{subcaption}

\usepackage[pagebackref=true,breaklinks=true,colorlinks,bookmarks=false]{hyperref}

\cvprfinalcopy 

\def\cvprPaperID{****} 

\ifcvprfinal\fi
\begin{document}

\title{GlamTry: Advancing Virtual Try-On for High-End Accessories}

\author{Ting-Yu Chang\\
Civil and Environmental Engineering\\
Stanford University\\
{\tt\small tingyuc@stanford.edu}
\and
Seretsi Khabane Lekena\\
Stanford Center for Professional Dvelopment\\
Stanford University\\
{\tt\small sklekena@stanford.edu}
}
\maketitle

\begin{abstract}
   The paper aims to address the lack of photorealistic virtual try-on models for accessories such as jewelry and watches, which are particularly relevant for online retail applications. While existing virtual try-on models focus primarily on clothing items, there is a gap in the market for accessories. This research explores the application of techniques from 2D virtual try-on models for clothing, such as VITON-HD, and integrates them with other computer vision models, notably MediaPipe Hand Landmarker. Drawing on existing literature, the study customizes and retrains a unique model using accessory-specific data and network architecture modifications to assess the feasibility of extending virtual try-on technology to accessories. Results demonstrate improved location prediction compared to the original model for clothes, even with a small dataset. This underscores the model's potential with larger datasets exceeding 10,000 images, paving the way for future research in virtual accessory try-on applications.
\end{abstract}

\section{Introduction}

\subsection{Literature Review}
Jeongho Kim et al. introduce StableVITON \cite{kim2023stableviton}, an approach aimed at enhancing the applicability of pre-trained diffusion models for image-based virtual try-on tasks. The primary focus is on preserving clothing details while leveraging the robust generative capabilities of these models. StableVITON achieves this by learning semantic correspondence between clothing and human body within the latent space of the pre-trained diffusion model. Notably, the proposed zero cross-attention blocks maintain clothing details and generate high-fidelity images by incorporating the pre-trained model's inherent knowledge during the warping process. Furthermore, the introduction of a novel attention total variation loss and augmentation techniques enhances the precision of clothing details representation. Through qualitative and quantitative evaluations, StableVITON demonstrates superior performance over baselines, showcasing promising results across diverse images.

Similar to StableVITON, IDM-VTON \cite{choi2024improving} proposed by Yisol Choi et al. is also a novel diffusion model tailored for image-based virtual try-on tasks. Unlike previous approaches, IDM-VTON effectively preserves garment identity by leveraging two modules to encode garment semantics: 1) high-level semantics integrated into the cross-attention layer, and 2) low-level features incorporated into the self-attention layer. Additionally, detailed textual prompts for both garment and person images enhance the authenticity of the generated visuals. Finally, IDM-VTON presents a customization method using a pair of person-garment images, significantly improving fidelity and authenticity. Experimental results show IDM-VTON outperforming StableVITON and other methods, demonstrating superior fidelity and authenticity in virtual try-on images in a real-world scenario.

In addition, both StableVITON and IDM-VTON utilize VITON-HD \cite{choi2021viton} as the training dataset to train their respective models. The availability of high-resolution data from VITON-HD is crucial for training robust virtual try-on models capable of synthesizing high-quality images. By leveraging the detailed information provided by VITON-HD, StableVITON and IDM-VTON are able to learn accurate semantic correspondences between clothing items and human bodies, resulting in improved fidelity and authenticity in the generated virtual try-on images. This highlights the importance of high-quality training data like VITON-HD in advancing the state-of-the-art in virtual try-on technology. In this research, we will employ a similar process to create a dataset for accessories. By collecting high-resolution images featuring individuals wearing various accessories such as rings and watches, we aim to compile a diverse dataset that encompasses a wide range of accessory types, poses, lighting conditions, and backgrounds. This dataset will serve as a valuable resource for training and evaluating virtual try-on models tailored specifically for accessories, allowing us to advance the state-of-the-art in this domain.


\section{Dataset}
In the VITON model family, VITON-HD\cite{choi2021viton}, StableVITON\cite{kim2023stableviton}, and  IDM-VTON\cite{choi2024improving} all utilize the VITON-HD dataset for training. Seunghwan Choi et al.  collected a high-resolution virtual try-on dataset, consisting of 13,679 pairs of frontal-view woman and top clothing images obtained from an online shopping mall website. The dataset was split into a training set with 11,647 pairs and a test set with 2,032 pairs. However, the majority of images in the VITON-HD dataset only feature clothing items, and we were unable to find a publicly available dataset online that sufficiently covers humans wearing accessories. Therefore, this research aims to construct the first comprehensive dataset specifically focused on accessories. We believe that this initiative will greatly benefit future researchers interested in virtual try-on applications for accessories.

\subsection{Data Collection}
\subsubsection{Web Scrapping}
To gather new data necessary for training the models towards our objective we built a filtering web scraper using available Google Vision and Search APIs. The first step involves accepting queries to gather images from Google Images and the second step filters them using the google Vision API to remove outliers that do not meet our criteria. Input queries are important here to ensure we gather diverse but relevant images to reduce manual cherry-picking later.
\subsubsection{Kaggle}
For the try-on items images, we utilized two datasets available on Kaggle. The watch dataset \cite{watch_data} comprises 2000 watch images, categorized into 5 different watch brands. On the other hand, the Tanishq Jewellery Dataset \cite{ring_data} includes images featuring two primary types of accessories: rings and necklaces. 
\subsection{Data Pre-Processing}
\begin{figure}[ht]
\begin{center}
\includegraphics[width=0.9\linewidth]{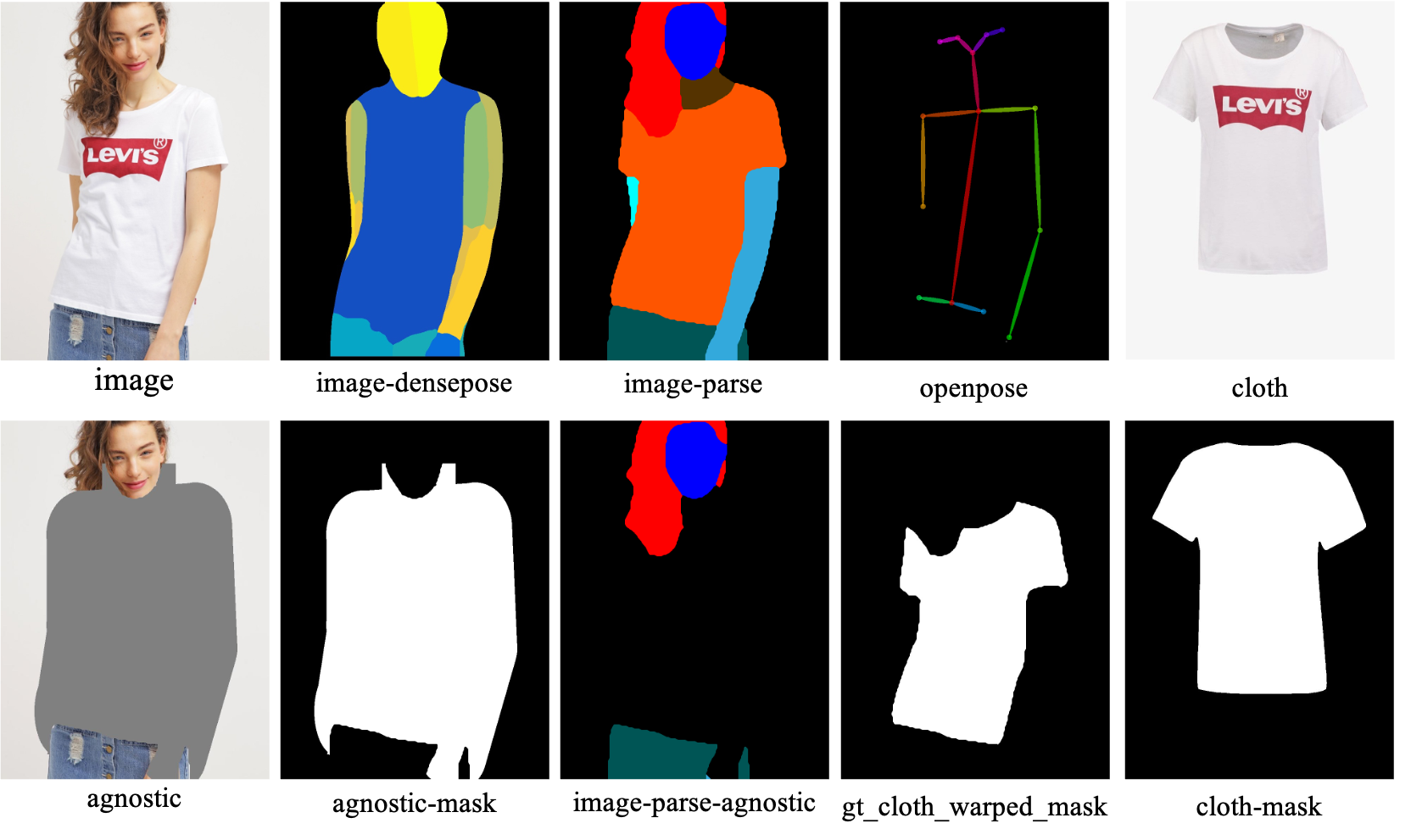}
\end{center}
   \caption{Overview of VITON-HD dataset.}
\label{fig:VITON-HD}
\end{figure}

After collecting the raw data, we follow the VITON-HD data pre-processing steps. Fig.\ref{fig:VITON-HD} provides an overview of all data types in the VITON-HD dataset. For the StableVITON model, we require agnostic, agnostic-mask, and image-densepose. To obtain these images, we need to complete four steps, including (1) Dense-pose, (2) Human Parsing\cite{li2020self} (3) Pose-estimation\cite{8765346}\cite{simon2017hand}\cite{cao2017realtime}\cite{wei2016cpm}, (4) Accessories-mask, (5)agnostic-mask and (6) human-agnostic. These steps allow us to construct our three target images.

\subsubsection{Human Parsing}
\begin{figure}[ht]
\begin{center}
\includegraphics[width=0.9\linewidth]{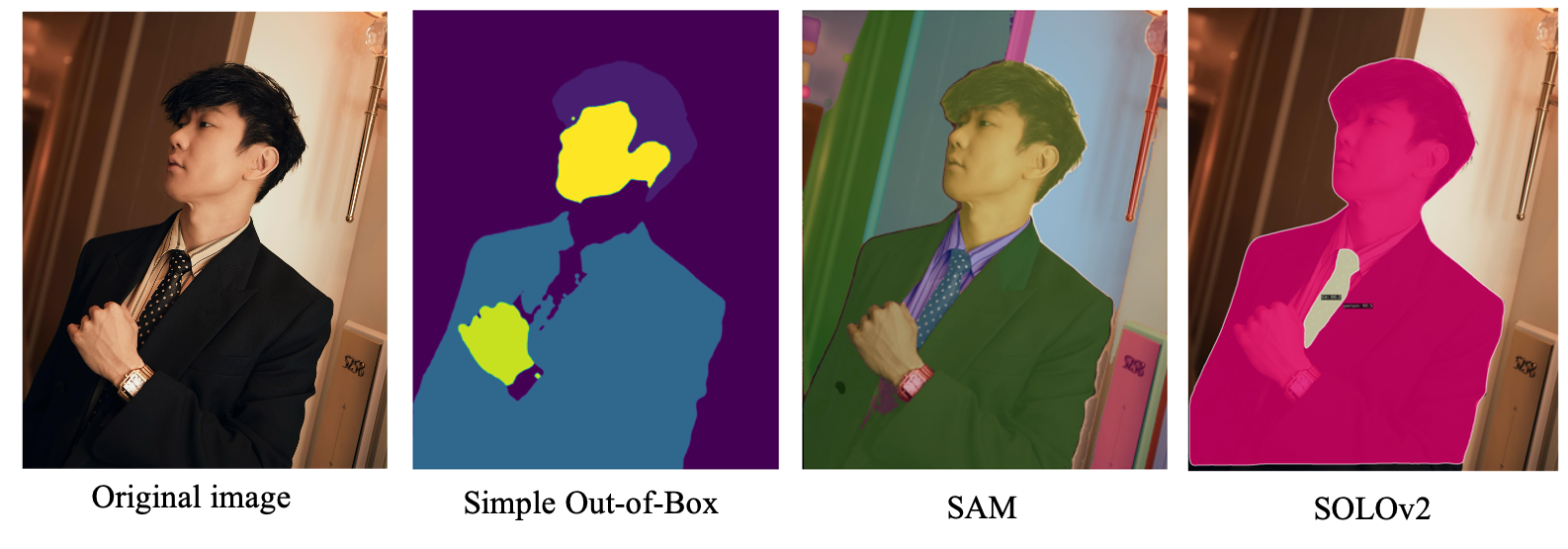}
\end{center}
\caption{The results of human parsing obtained from different models.}
\label{fig:Human Parse}
\end{figure}

We experimented with three different models to obtain the Human Parsing feature for the agnostic-mask: (1) Simple Out-of-Box Extractor trained with Look into Person (LIP) dataset \cite{li2020self}, (2) Segment Anything (SAM) \cite{kirillov2023segany}, and (3) SOLOv2 from MMdetection \cite{wang2020solov2}. LIP is the largest single person human parsing dataset with 50000+ images. This dataset focus more on the complicated real scenarios. LIP has 20 labels, including 'Background', 'Hat', 'Hair', 'Glove', 'Sunglasses', 'Upper-clothes', 'Dress', 'Coat', 'Socks', 'Pants', 'Jumpsuits', 'Scarf', 'Skirt', 'Face', 'Left-arm', 'Right-arm', 'Left-leg', 'Right-leg', 'Left-shoe', 'Right-shoe'. In Fig.\ref{fig:Human Parse}, we observed that the Simple Out-of-Box Extractor performs the best in segmenting the watch accurately. However, the watch was classified as background due to the absence of watches in the training labels. SAM also provides accurate segmentation results, but the mask lacks a specific label. In contrast, SOLOv2 categorizes the watch as part of the human.

To address the problem, we plan to retrain the Simple Out-of-Box Extractor with a parsing dataset that includes 12,701 images with manually annotated parsing labels from DeepFashion-MultiModal \cite{jiang2022text2human}. The parsing dataset comprises 24 categories, including desirable labels such as ring, wrist wearing, earrings, and necklace. We aim to improve the segmentation results by incorporating this dataset into the training process.

\subsubsection{OpenPose}
OpenPose \cite{8765346} pioneered the first real-time multi-person system capable of simultaneously detecting human body, hand, facial, and foot key points in single images, such as \{"Nose": 0, "Neck": 1, "RShoulder": 2, "RElbow": 3, "RWrist": 4, "LShoulder": 5, "LElbow": 6, "LWrist": 7, "RHip": 8, "RKnee": 9, "RAnkle": 10, "LHip": 11, "LKnee": 12, "LAnkle": 13, "REye": 14, "LEye": 15, "REar": 16, "LEar": 17, "Background": 18\}, totaling 19 key points. Expanding the data points of OpenPose to make use of hand pose estimation did not improve our data set here. Wrist estimations accuracy did not improve against our dataset \cite{8765346, simon2017hand, cao2017realtime, wei2016cpm}. We utilize pose estimation features along with human parsing to pre-process human images, transforming them into human-agnostic images for use in the final query data. 

As depicted in Fig.\ref{fig:Feature} (b), we observe that the model still predicts key points for the leg or lower body. However, our images primarily focus on the body part above the waist, as we are primarily interested in the upper body or even hand features for accessories. Thus, further data processing may be required, or we can choose to ignore the key points below the waist.

\subsubsection{MediaPipe}
\begin{figure}[ht]
\begin{center}
\includegraphics[width=0.9\linewidth]{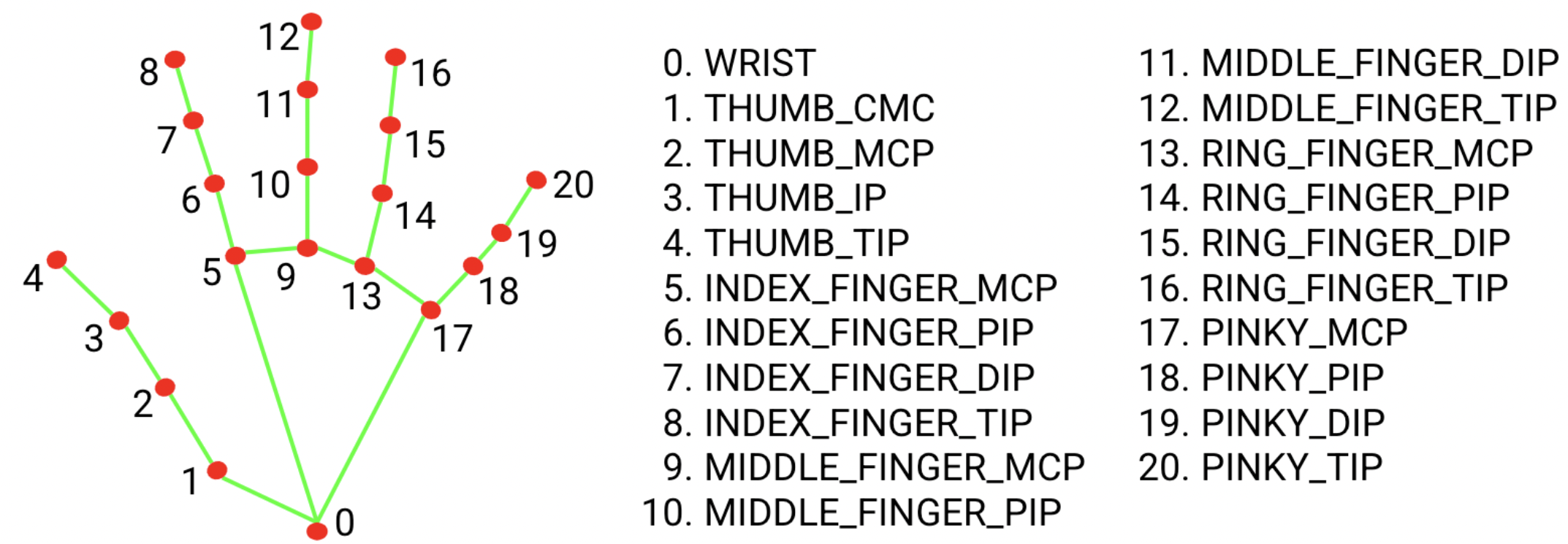}
\end{center}
\caption{The predicted 21 key points of MediaPipe Hand Landmarker.}
\label{fig:HM}
\end{figure}

To enhance the detection of hand features, we incorporate the MediaPipe Hand Landmarker task \cite{zhang2020mediapipe}. This model accurately identifies and localizes 21 key points on the hand (See Fig.\ref{fig:HM}), covering all knuckle coordinates within the detected hand regions. It has been trained on a diverse dataset of approximately 30K real-world images and numerous synthetic hand models superimposed on various backgrounds. Fig.\ref{fig:Feature} (c) shows the results visualization of the hand pose estimation. We select keypoints $0, 9,$ and $13$ to develop a straightforward algorithm for predicting the watch location. In addition, we save the hand pose estimation as an image to be used as an input for the synthesis generator (See the right image of Fig.\ref{fig:Feature} (c)).

\subsubsection{Accessories-mask}
\begin{figure}[ht]
\begin{center}
\includegraphics[width=1.1\linewidth]{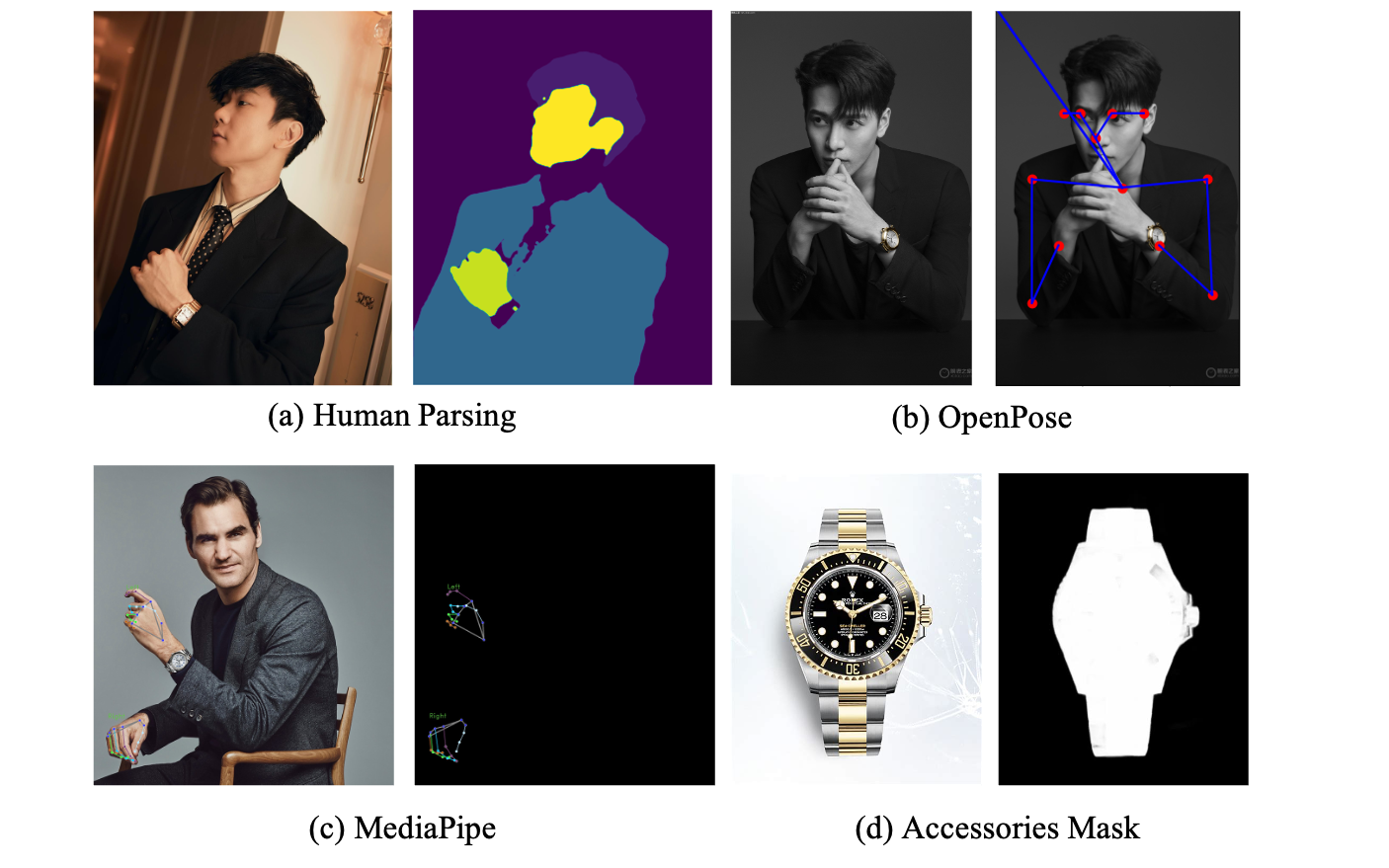}
\end{center}
\caption{(a) The original image of people wearing watch and its human Parsing output. (b) The original image of people wearing watches and its OpenPose output. (c) The original image of people wearing watches and its MediaPipe output. (d) The original image of the target watch from Kaggle and its mask.}
\label{fig:Feature}
\end{figure}

The accessories mask of the items we are interested in trying on represents the agnostic mask of one of the inputs for the pre-trained U-Net, which serves as the query (Q). We apply U2-net \cite{qin2020u2net} to obtain the outline of the items. Fig.\ref{fig:Feature} (d) illustrates the mask result of the target watch for the try-on process.

\subsubsection{Agnostic-mask and Human-agnostic}
Agnostic-mask and Human-agnostic are crucial images for the model's input and can be generated after preparing the previous data types. Our goal is to segment the target accessory mask in human parsing using OpenPose estimation. Specifically, we focus on the labels "RWrist" and "LWrist" of the keypoints stored in the JSON file obtained from OpenPose, as well as the "background" label of the human-parsing mask since the watch is classified under this label. 

\begin{figure}[ht]
\begin{center}
\includegraphics[width=1.1\linewidth]{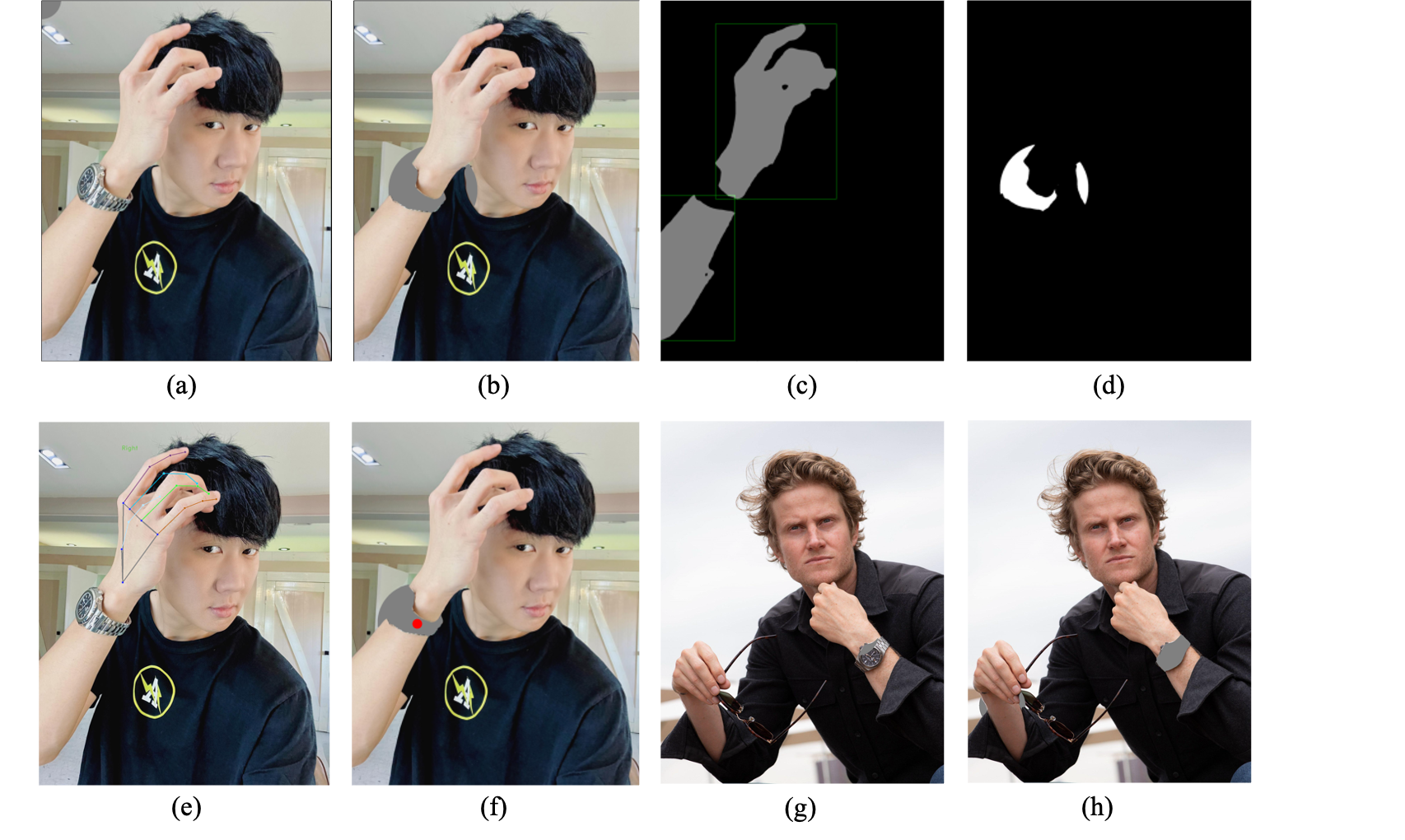}
\end{center}
\caption{(a) The initial unsuccessful prediction of the human-agnostic mask. (b) The enhanced human-agnostic mask achieved using our algorithm. (c) Arm masks with bounding boxes overlaid. (d) Agnostic mask highlighting the targeted watch region. (e) The original image with predicted MediaPipe Hand Landmarker hand pose. (f) The predicted watch location using the midpoint algorithm. (g) An instance of unsuccessful watch mask prediction. (h) An improved watch mask prediction by the midpoint algorithm.}
\label{fig:agnostic}
\end{figure}

First, we establish a threshold for the distance between the wrists and create a circular mask with the wrist points as the center. Then, we perform an intersection operation between the background mask and the circle to obtain the desired region. While this approach yields satisfactory results, we observed instances where individuals are positioned close to the camera, resulting in incomplete wrist point predictions, denoted as $(x_{\text{wrist}}, y_{\text{wrist}}) = (0,0)$. Consequently, the generated mask appears as a circle in the corner (See Fig.\ref{fig:agnostic} (a)). 

To address this issue, we developed an algorithm that relies solely on the parsing mask to generate the watch mask. Initially, we combine the "Left-arm" and "Right-arm" masks and identify the two largest contours, which should correspond to the same labeled mask separated by the watch. Subsequently, we calculate the bounding boxes (See Fig.\ref{fig:agnostic} (c)) of the two contours, along with the coordinates of their centroids. The midpoint between the centroids serves as the estimated location for the missing wrist keypoints. After acquiring the agnostic-mask, we apply it to the original image by overlaying it in gray color. Fig.\ref{fig:agnostic} (b) demonstrates a significant improvement in performance achieved with our heuristic algorithm.

However, despite the improvement, some of the predicted parsing watch masks remain inaccurate, as shown in Fig.\ref{fig:agnostic} (g). To address this, we utilize the MediaPipe Hand Landmarker task, as mentioned in the previous section, in an effort to achieve better results. We begin by finding the midpoint between $Keypoint\ 9$ and $Keypoint\ 13$, and then treat $Keypoint\ 0$ as the midpoint between this midpoint and the watch location. The red dot in the Fig.\ref{fig:agnostic} (f) is the predicted watch location. This approach is necessary because the wrist keypoint provided by MediaPipe does not precisely correspond to the actual location of the watch (See Fig.\ref{fig:agnostic} (e)). Furthermore, we identified an issue with the heuristic algorithm: it uses a fixed radius for the circle, which is not suitable given the varying image sizes. To address this, we calculate and store the distance between the predicted watch location and $Keypoint\ 0$ as the radius for the circle mask. This radius is then used to create the circle, and we perform an intersection with the parse image to refine the mask. Fig.\ref{fig:agnostic} (h) illustrates an example of an improved result using the revised algorithm.
\section{Method}
\begin{figure}[ht]
\begin{center}
\includegraphics[width=1.1\linewidth]{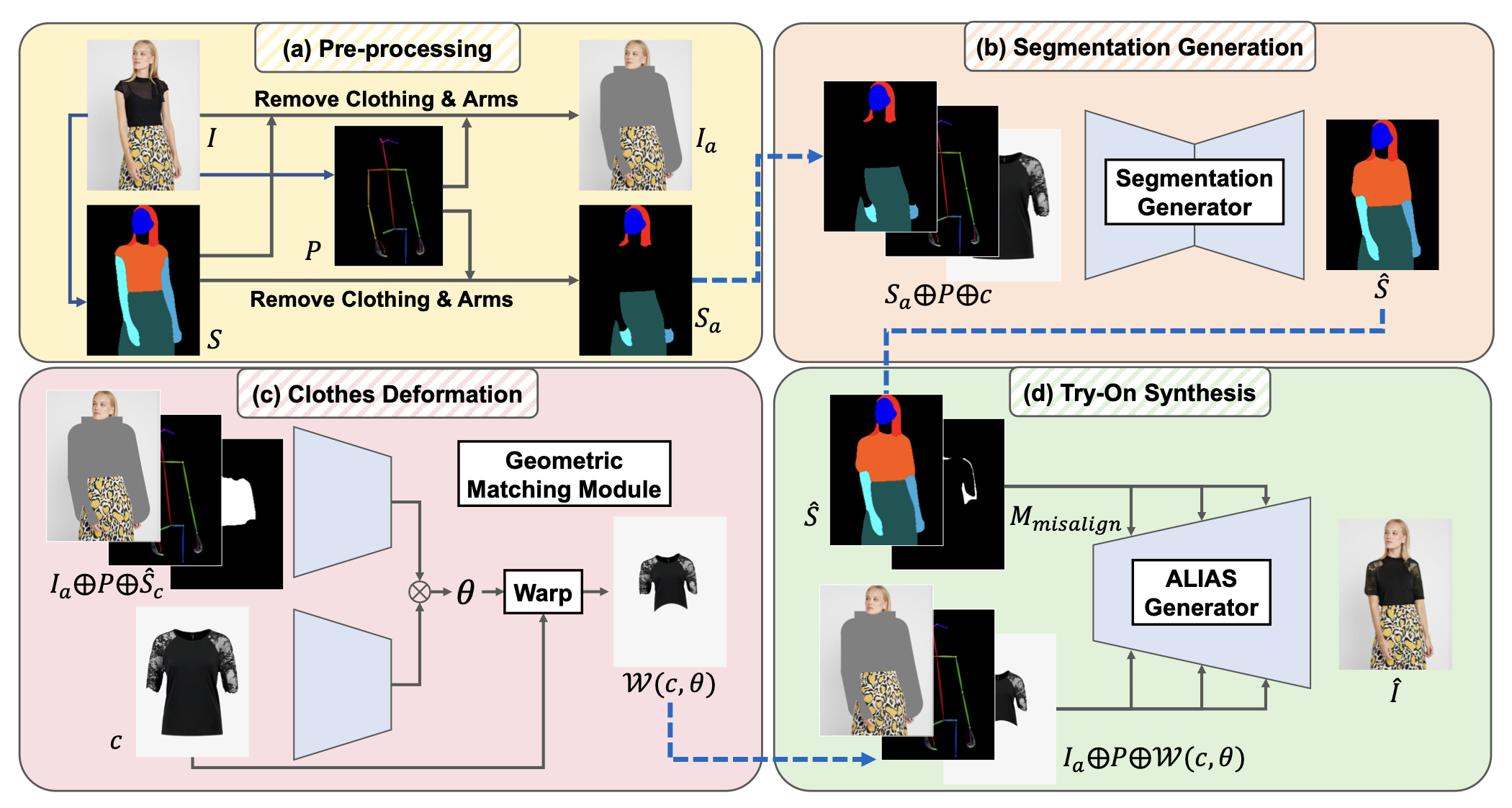}
\end{center}
   \caption{Overview of a VITON-HD. (a) First, given a reference image $I$ containing a target person, we predict the segmentation map $S$ and the pose map $P$, and utilize them to pre-process $I$ and $S$ as a clothing-agnostic person image $I_a$ and segmentation $S_a$. (b) Segmentation generator produces the synthetic segmentation ${\hat{S}}$ from ($S_a$,$P$,$c$). (c) Geometric matching module deforms the clothing image $c$ according to the predicted clothing segmentation ${\hat{S_c}}$ extracted from ${\hat{S}}$. (d) Finally, ALIAS generator synthesizes the final output image ${\hat{I}}$ based on the outputs from the previous stages via our ALIAS normalization.\cite{choi2021viton}}
\label{fig:HD}
\end{figure}

Our model architecture is primarily based on VITON-HD \cite{choi2021viton}. An overview of the VITON-HD model is presented in Fig. \ref{fig:HD}. We skip the segmentation part by utilizing our predicted watch mask. Since we aim to assess the ability to extend the clothing try-on task to watches within the pre-trained VITON-HD framework, we employ the pre-trained Geometric Matching Module model and ALIAS Generator with the data following the pre-processing steps outlined previously as the baseline model to evaluate its performance on watches. Here, we maintain the pose map $P$ as the pose estimation obtained by OpenPose to align with VITON-HD, instead of using the proposed MediaPipe as a baseline.

\begin{figure}[ht]
\begin{center}
\includegraphics[width=1.05\linewidth]{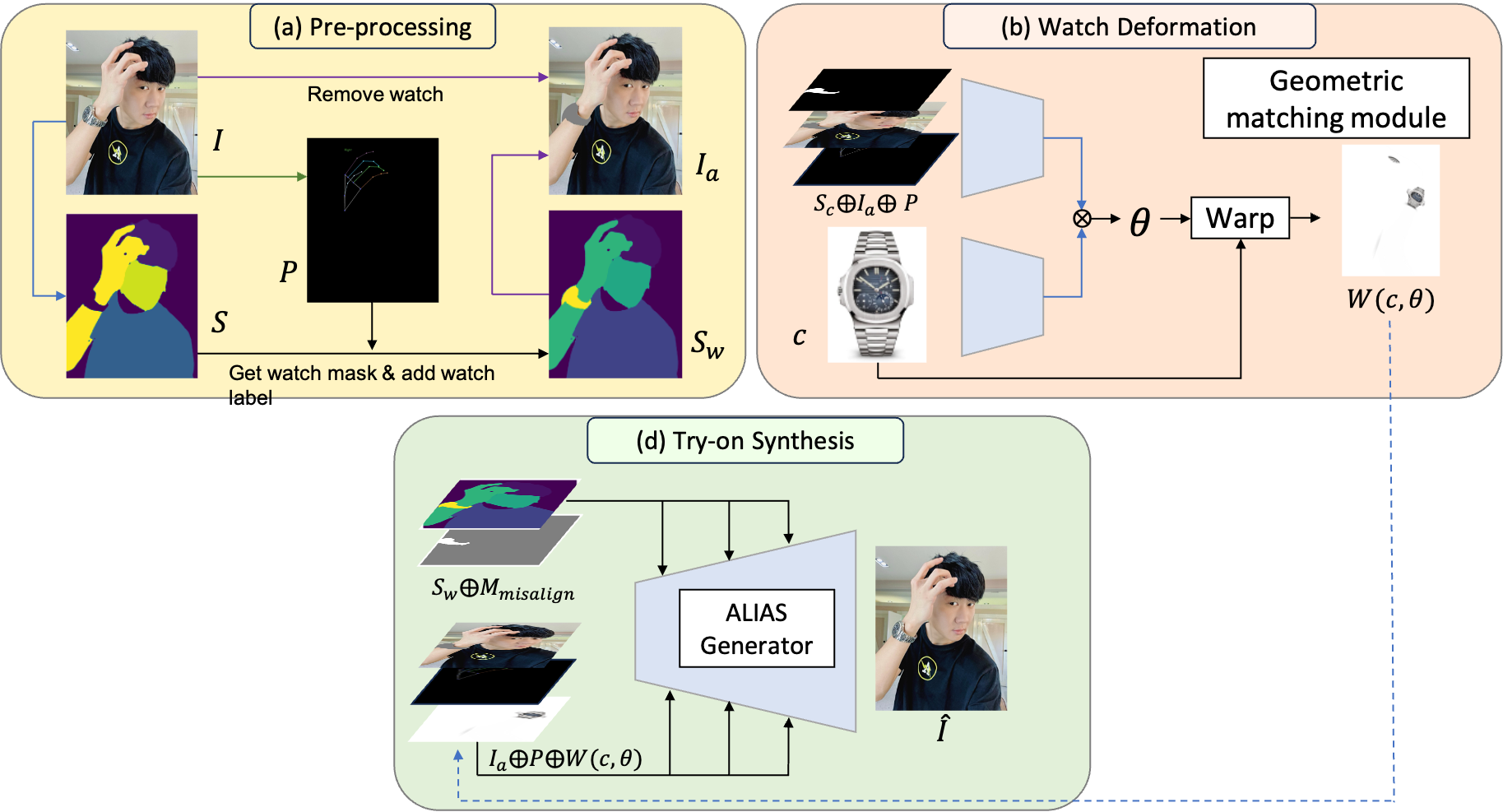}
\end{center}
   \caption{Our model architecture designed to test the hypothesis for achieving improved results.}
\label{fig:method}
\end{figure}

The baseline results are expected to be poor for two main reasons. Firstly, the entire human pose may not be directly relevant to the position of the watch. Secondly, the model is trained primarily for clothing, meaning that the clothes deformation model may not accurately align the watch with the correct position. Therefore, we need to replace the OpenPose pose map with the output from the MediaPipe Hand Landmarker and subsequently retrain the Geometric Matching Module (GMM) first proposed in CP-VTON \cite{wang2018toward} using our watch data. Fig.\ref{fig:method} represents our model architecture designed to test our hypothesis for achieving improved results.

\begin{figure}[ht]
\begin{center}
\includegraphics[width=1\linewidth]{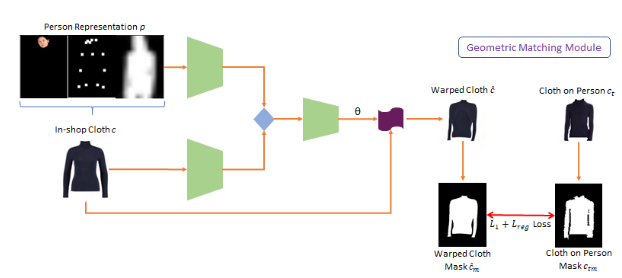}
\end{center}
   \caption{Full GMM pipeline of CP-VTON+.}
\label{fig:CP-VTON}
\end{figure}

\begin{figure}[ht]
\begin{center}
\includegraphics[width=1.05\linewidth]{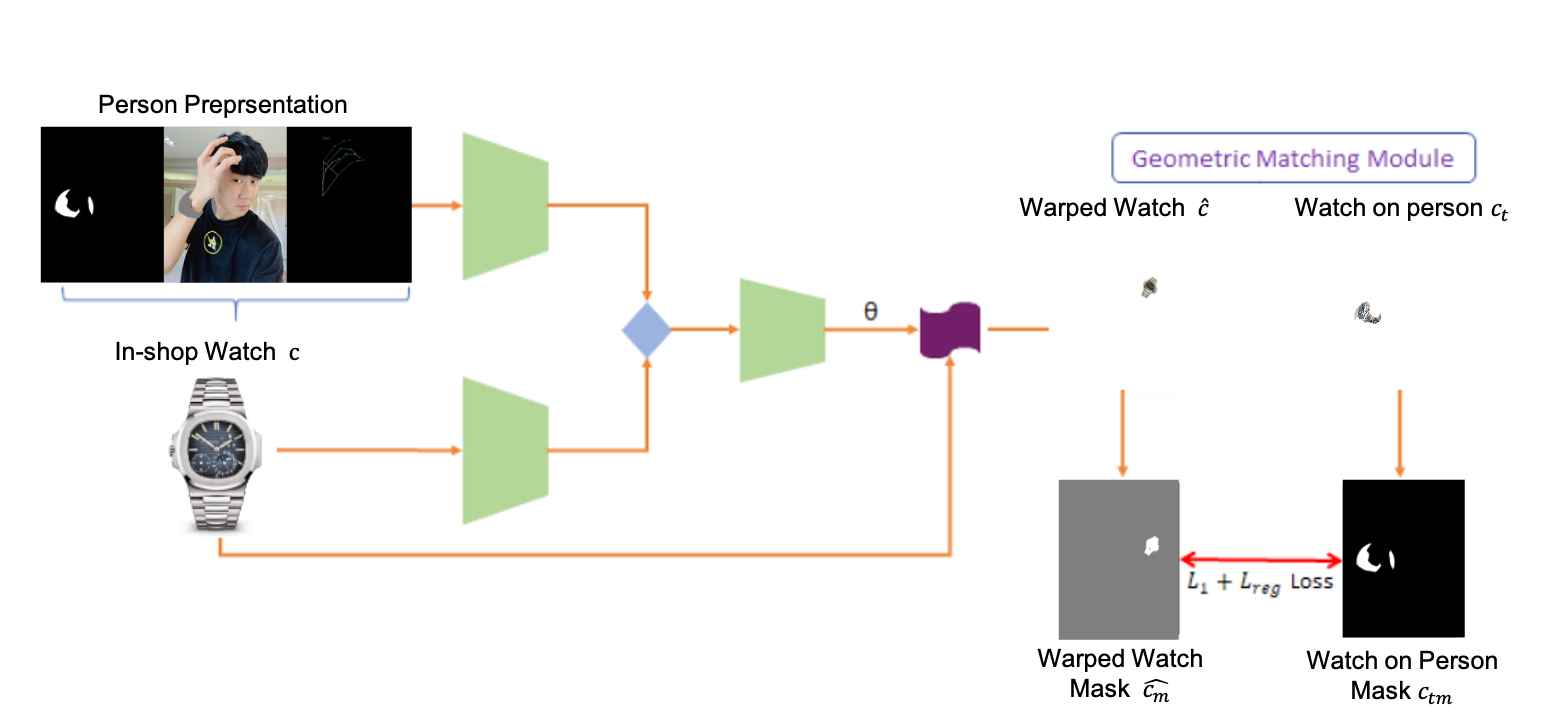}
\end{center}
   \caption{Full GMM pipeline for our task.}
\label{fig:Our CP-VTON}
\end{figure}

CP-VTON adopted a geometric matching module to learn the parameters of TPS transformation, which improves the accuracy of deformation. Here, we customize CP-VTON+ \cite{minar2020cpvton} architecture for our task. CP-VTON+ is named after the baseline CP-VTON, outperforms CP-VTON by large margins, in both perceptible and subjective evaluations. The CP-VTON GMM network is built on CNN geometric matching. Whereas the CNN geometric matching uses a pair of color images, CP-VTON GMM inputs are binary mask information, silhouette, and joint heatmap and the colored try-on clothing (See Fig.\ref{fig:CP-VTON}). Here we follow VITON-HD to use binary mask of watch, hand pose map and colored try-on clothing (See Fig.\ref{fig:Our CP-VTON}). 

This GMM (Geometric Matching Module) network, is designed for geometric matching tasks, particularly for aligning two input images spatially. The network consists of several components. First, it employs two feature extraction modules (extractionA and extractionB) to extract high-level features from the input images. These features are then normalized using L2 normalization. The normalized features are passed through a correlation layer, which calculates the correlation between the features from the two images. This correlation represents the similarity between corresponding spatial locations in the two sets of images. The network then utilizes a regression module to predict the parameters of a thin-plate spline (TPS) transformation based on this correlation. These parameters are used to generate a dense grid of control points. Finally, a TPS grid generator module utilizes these control points to generate a transformation grid, which can be applied to one of the input images to align it with the other image.

As for the loss function, The GicLoss module is used to enforce geometric consistency between neighboring grid points in the generated transformation grid. This loss penalizes differences in distances between neighboring grid points before and after transformation, ensuring smooth and consistent deformation. Overall, the network and loss function work together to learn a transformation that aligns the input images both geometrically and perceptually.

\begin{equation} \label{eq:theta}
\theta = f_{\theta}\left(f_H(H_t), f_C({Ci})\right)
\end{equation}

\begin{equation} \label{eq:GMM}
L^{CP-VTON+}_{GMM} = \lambda_1 \cdot L_1(C_{\text{warped}}, I_{Ct}) + \lambda_{\text{reg}} \cdot L_{\text{reg}}
\end{equation}

\begin{equation} \label{eq:reg}
\begin{split}
L_{\text{reg}}(G_x, G_y) = & \sum_{i=-1,1} \sum_{x} \sum_{y} \left| G_x(x+i, y) - G_x(x, y) \right| \\
& + \sum_{j=-1,1} \sum_{x} \sum_{y} \left| G_x(x, y+j) - G_x(x, y) \right|
\end{split}
\end{equation}

In the CP-VTON+ experiments, it was observed that the clothing warping often resulted in significant distortion when compared to existing methods. While the exact cause remained unclear, it was evident that regularization of TPS parameters was necessary to account for the intricacies of clothing textures. To address this, the authors introduced grid warping regularization, focusing on grid deformation rather than directly manipulating TPS parameters. This approach was chosen for its simplicity and ease of visualization, aiming to minimize abrupt changes in warping between adjacent grid intervals, as depicted in equation \ref{eq:reg}.

\begin{figure}[ht]
\begin{center}
\includegraphics[width=1.03\linewidth]{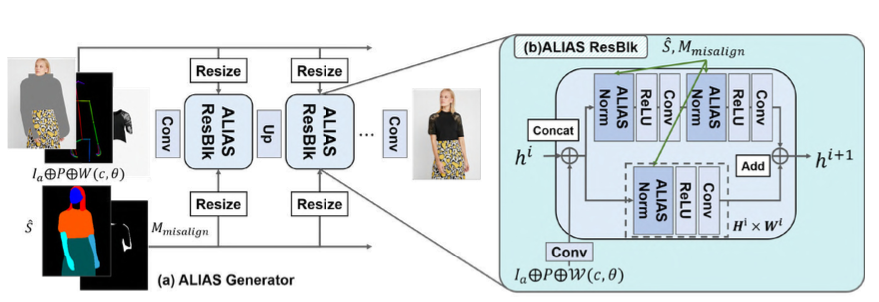}
\end{center}
   \caption{ALIAS generator. (a) The ALIAS generator is composed of a series of ALIAS residual blocks, along with up- sampling layers. The input ($I_a$, $P$, $W(c, \theta)$) is resized and injected into each layer of the generator. (b) A detailed view of a ALIAS residual block. Resized ($I_a$, $P$, $W(c, \theta)$) is concatenated to $h^i$ after passing through a convolution layer. Each ALIAS normalization layer leverages resized $\hat{S}$ and $M_{misalign}$ to normalize the activation. \cite{choi2021viton}}
\label{fig:ALIAS}
\end{figure}

We have decided to utilize the pre-trained ALIAS Generator with 100.5 million parameters in VITON-HD for both the baseline model and our proposed architecture, with a minor modification involving the replacement of the OpenPose map with MediaPipe in the latter. This decision stems from our belief that a well-trained model like ALIAS should be capable of embedding the extracted image features and generating high-quality results based on them, irrespective of variations such as different individuals, clothing items, or watches. One aspect we may question is whether ALIAS has fully grasped the nuanced relationship within $P$ solely through the entire human pose, potentially leading to inaccuracies in location alignment prediction. Fig. \ref{fig:ALIAS} shows the details of ALIAS Generator in VITON-HD.

\section{Experiment}
\subsection{Baseline Model}
\begin{figure}[ht]
\begin{center}
\includegraphics[width=1.1\linewidth]{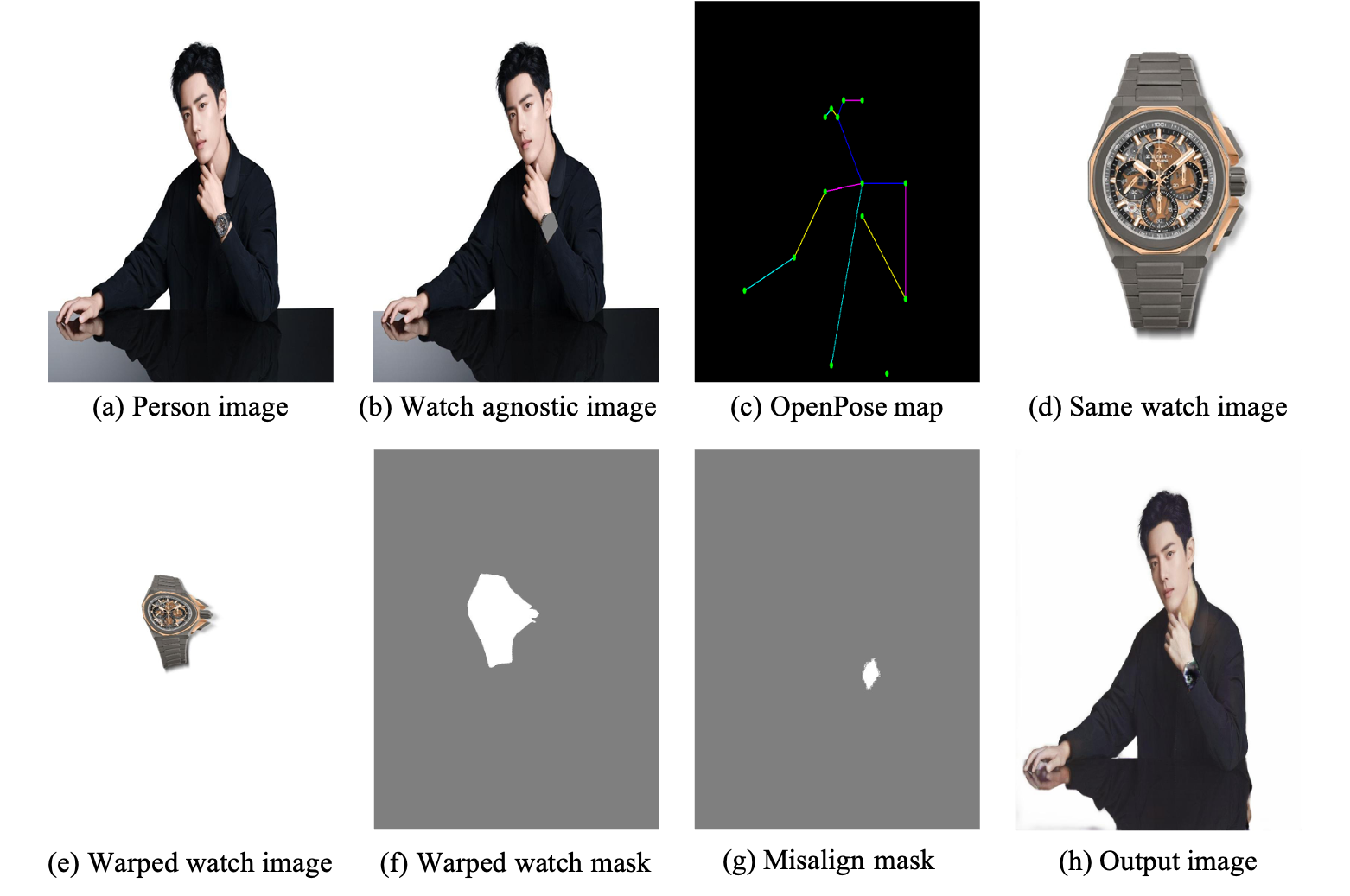}
\end{center}
   \caption{Experiment using the baseline model.}
\label{fig:baseline}
\end{figure}

Under the architecture of our method as depicted in Fig.\ref{fig:method}, but utilizing the OpenPose map along with the pre-trained GMM and ALIAS in VITON-HD, we obtain the generated images, warp image, and misalign mask as baseline results. Figure \ref{fig:baseline} (a), (b), (c), and (d) display the input images for one instance. We create a small test dataset from the images we collected, consisting of paired images of individuals and the watches they are wearing, which the model has not previously seen. These images are resized to 1024×768. This approach benefits the evaluation process. We treat the person images as the ground truth data to compare the differences between these and the generated results. 

\begin{figure}[ht]
\begin{center}
\includegraphics[width=1.1\linewidth]{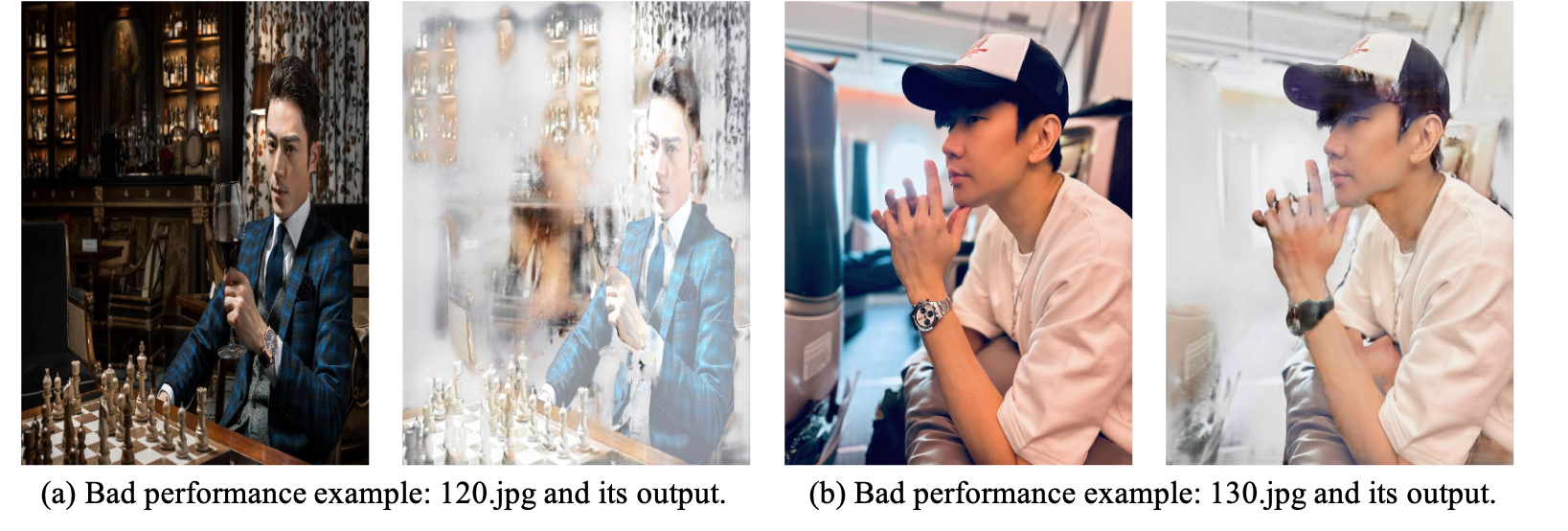}
\end{center}
   \caption{Poor output from the baseline model experiment.}
\label{fig:bad}
\end{figure}

Using the output of 96.jpg as an example, Figure \ref{fig:baseline} (e), (f), (g), and (h) illustrate that the warped watch image obtained by the GMM and the output from the ALIAS Generator in VITON-HD preserve good details of both the person and the watch. However, the location prediction does not perform as well as expected, resulting in the agnostic area not properly generating the target watch. Furthermore, we observe that images with non-white or complex backgrounds perform worse than this example. Figure \ref{fig:bad} (a) and (b) represent the results of images with complex backgrounds. While we can still discern the outline of the person and other objects, the colors are not preserved, resulting in faded images generated by the generator. We found that there are more instances like this than in the case of 96.jpg. We believe this is because the pre-trained ALIAS model is trained with uniform images featuring a white background; therefore, the model may struggle to effectively handle images with complex backgrounds. However, we are still impressed by the model's ability to preserve most of the image details despite the faded colors.

\subsection{Model with Watch Deformation}
\subsubsection{Retraining}
It's uncertain whether the issue we're observing in the baseline results stems from incorrect GMM predictions or from the ALIAS model itself. Therefore, we've decided to prioritize retraining the GMM model initially to assess if any improvements can be achieved.

Due to time constraints, we've compiled a small training dataset consisting of hundreds of images for GMM training within the allocated time frame. To monitor the training progress and maintain a record of the training history, we're using TensorBoard. We use the Adam optimizer to update the parameters of our model, with a specified learning rate (lr) set to opt.lr and momentum parameters (betas) of 0.5 and 0.999. Additionally, we implement a step-based learning rate decay strategy, where the learning rate is reduced every 10,000 steps, with the total number of steps before adjustment determined by $keep\_step$ + $decay\_step$.

\begin{figure}[ht]
\begin{center}
\includegraphics[width=0.85\linewidth]{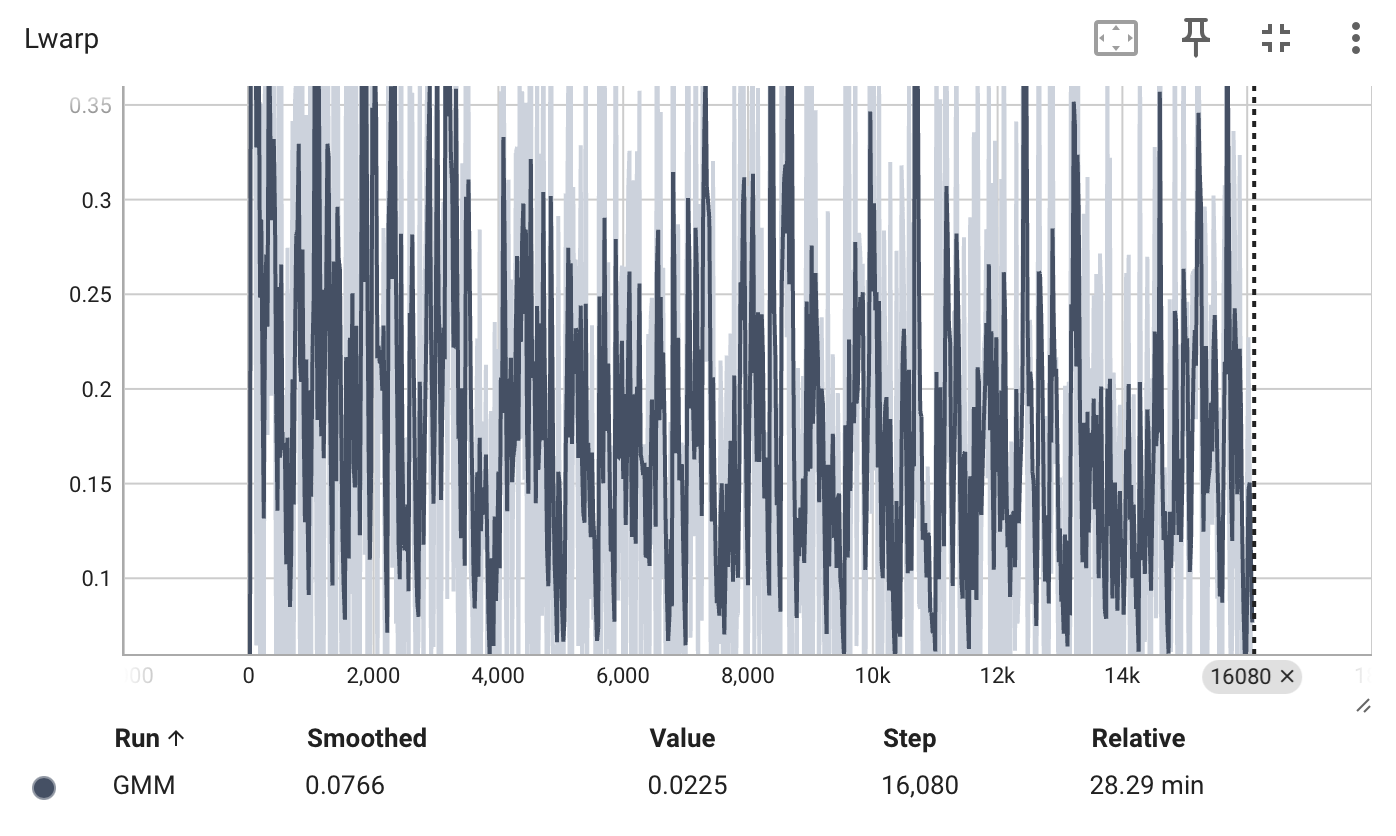}
\end{center}
   \caption{Training history of $\lambda_1 \cdot L_1$}
\label{fig:Lwarp}
\end{figure}

\begin{figure}[ht]
\begin{center}
\includegraphics[width=0.85\linewidth]{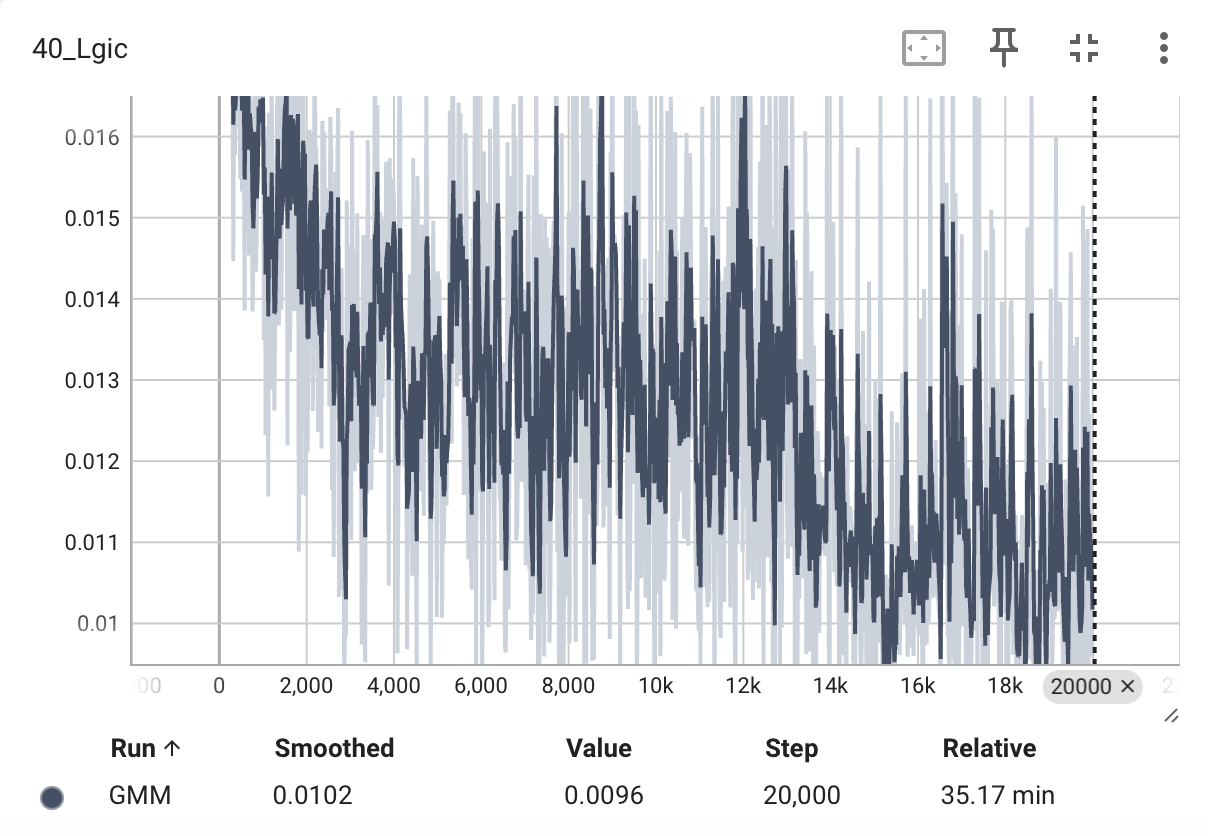}
\end{center}
   \caption{Training history of $\lambda_{reg} \cdot L_{reg}$}
\label{fig:Lgic}
\end{figure}

Fig.\ref{fig:Lwarp} and \ref{fig:Lgic} visualize the training history of first and second components of $L^{CP-VTON+}_{GMM}$ in equation \ref{eq:reg}. It's observed that the loss values for both components oscillate during training. Despite the oscillations, the overall trend of the loss values of $\lambda_{reg} \cdot L_{reg}$ demonstrate a gradual decrease over epochs. However, the loss values corresponding to \( \lambda_1 \cdot L_1 \) do not exhibit this desirable trend, suggesting the need for potential adjustments in the learning rate or fine-tuning of the weighted coefficient \( \lambda_1 \).

\begin{figure}[ht]
\begin{center}
\includegraphics[width=0.85\linewidth]{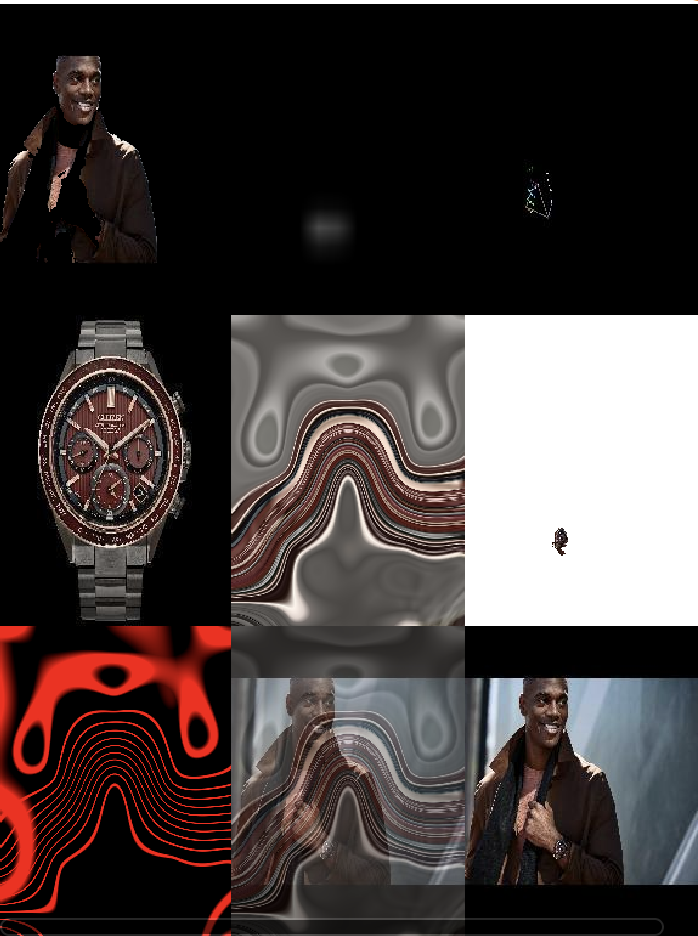}
\end{center}
   \caption{The severely distorted warped grid and image.}
\label{fig:distort}
\end{figure}

We've also encountered a challenge similar to that reported in previous studies, where some of the warped watches are severely distorted. As depicted in Fig.\ref{fig:distort}, a instance showcases this issue, where it's difficult to discern the watch in the center image, and the bottom-left warped image is notably distorted. The color of the background could be a contributing factor to this issue. We noticed that most of the watch images with a dark background exhibit this problem, whereas we haven't encountered it with images featuring a white background.

\section{Results/Evaluation}
\subsection{Results}
\begin{figure}[ht]
\begin{center}
\includegraphics[width=1\linewidth]{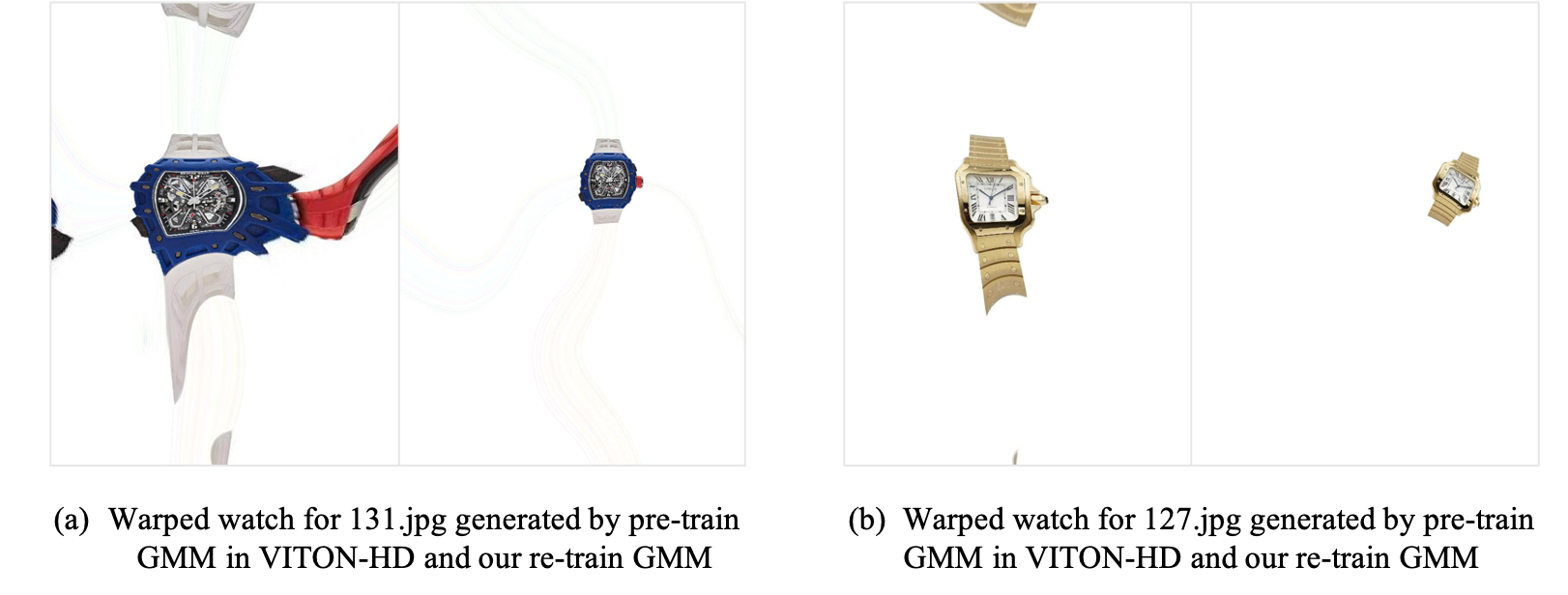}
\end{center}
   \caption{Comparison of the warped watches generated by pre-train GMM and our re-train GMM.}
\label{fig:warp_watch}
\end{figure}

\begin{figure}[ht]
\begin{center}
\includegraphics[width=0.85\linewidth]{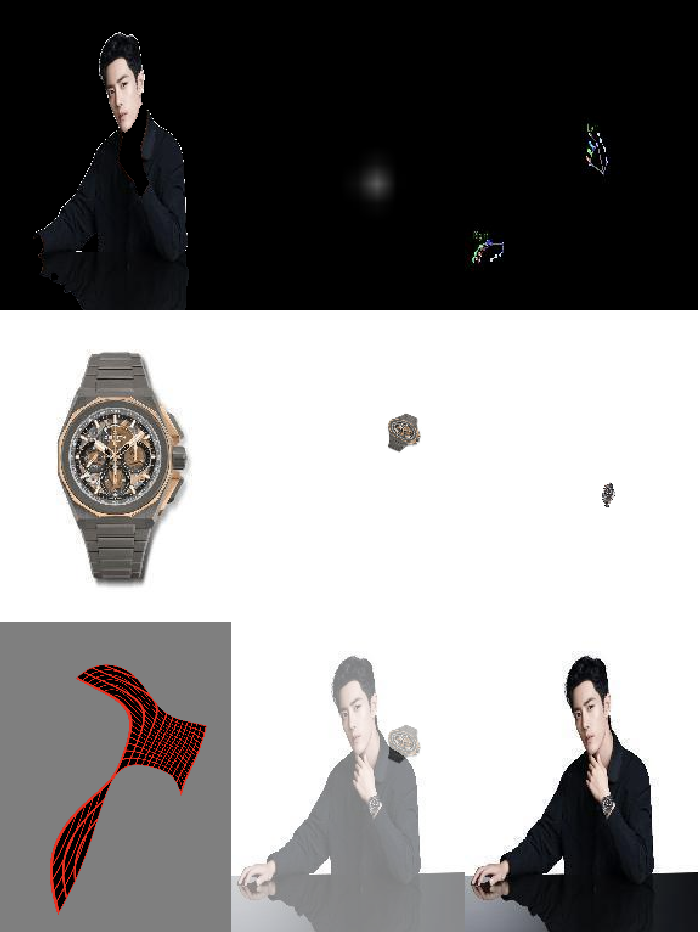}
\end{center}
   \caption{Demonstration of trend improvement using the example image 96.jpg.}
\label{fig:result}
\end{figure}

First, we compare the quality of warped watch generation between the pre-trained GMM in VITON-HD and our re-trained GMM. We observed that our model preserves the shape of the watch better (See Fig.\ref{fig:warp_watch}) and the size is more reasonable relative to the scale of the person in the image. Although the predicted location is still not accurate, we can see that it is closer to the target area compared to the baseline results. Taking 96.jpg as an example again, Fig.\ref{fig:result} illustrates the improvement in trend. We can anticipate this outcome given that we only have a small dataset for training compared to the GMM in VITON-HD, which is trained with approximately 16,000 images. Our dataset is insufficient for the model to learn the relative location relationship between the watch mask of the target watch and the watch-agnostic areas in the person image. This experiment has demonstrated improvement, suggesting further potential enhancement with a larger dataset.

\begin{figure}[ht]
\begin{center}
\includegraphics[width=0.85\linewidth]{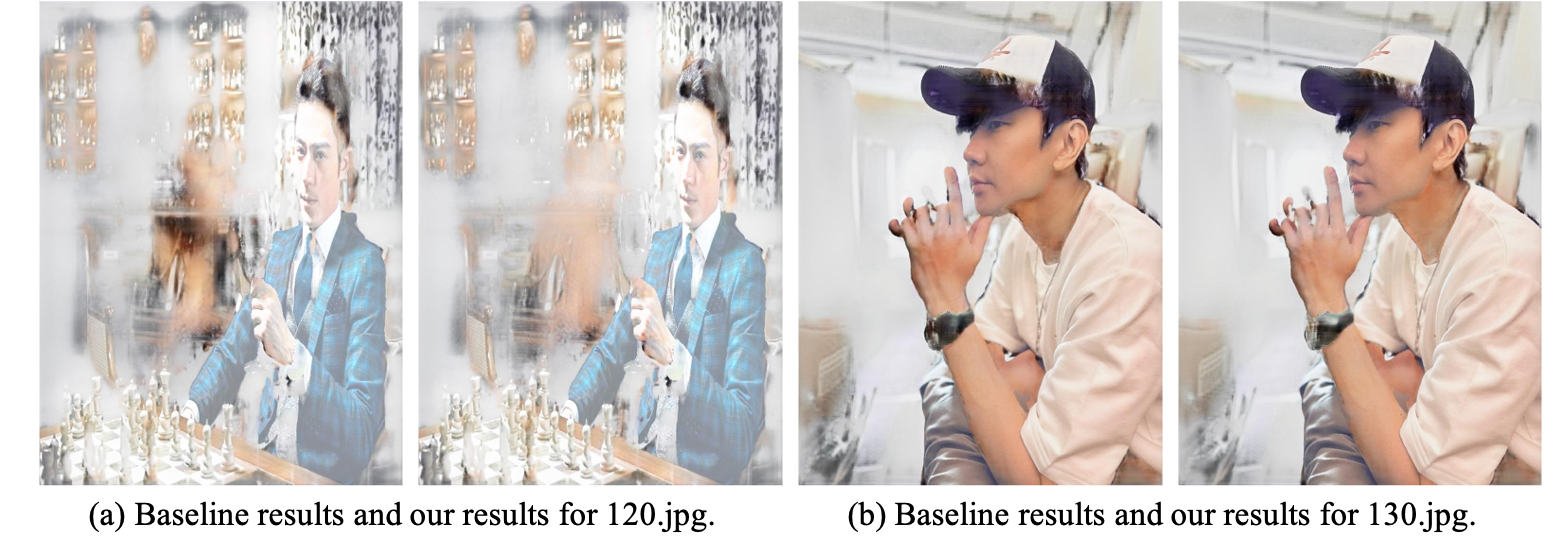}
\end{center}
   \caption{Comparison between the output of baseline model and ours.}
\label{fig:fad}
\end{figure}

The faded color issue was not resolved by retraining the GMM. Figure \ref{fig:fad} compares the baseline results with the proposed method, using the same examples as before. There is little difference between the baseline and the proposed method because the locating prediction is still inaccurate and the ALIAS Generator has not been retrained with the images with complex background.

\subsection{Qualitative Results}
\begin{figure}[ht]
    \centering
    \includegraphics[width=0.5\textwidth]{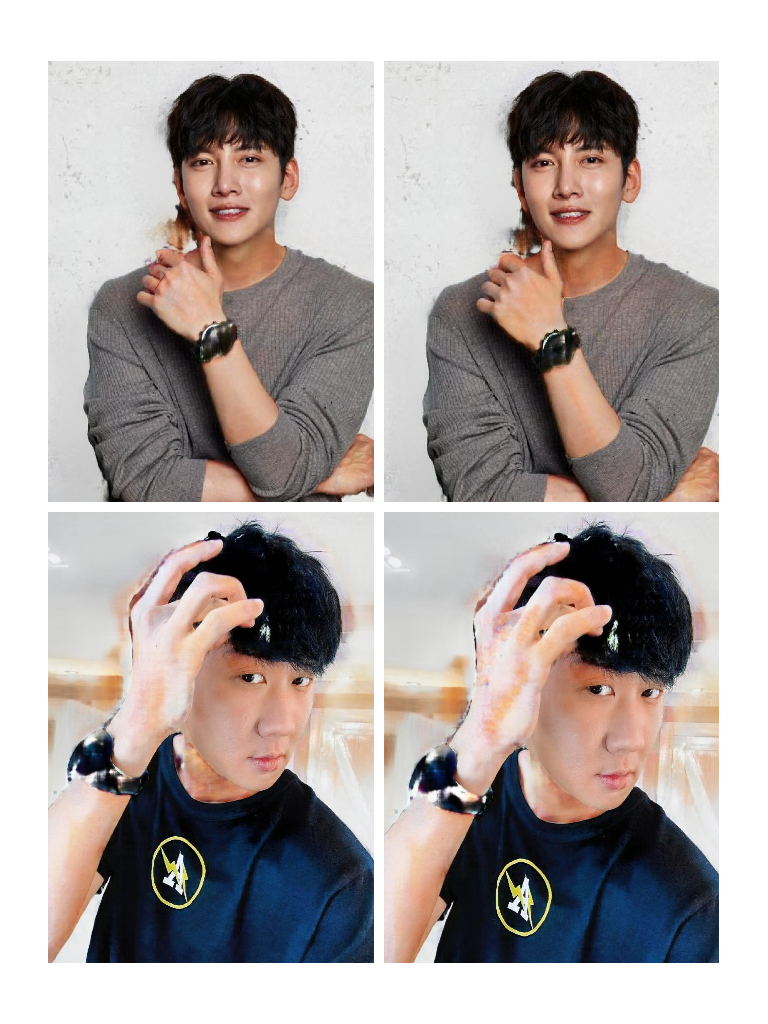}
    \caption{Qualitative side by side. While the images show strong distortion, ours (right two) shows less distortion on the rest of the body.}
    \label{fig:qual1}
\end{figure}

Since Virtual Try-On is a practical application of generative models, qualitative evaluation can be conducted directly by human observation. Performance can be assessed by comparing images generated by different models, figure \ref{fig:qual1}. Furthermore, we aim to create a simple questionnaire to gather feedback from our friends and classmates on the quality of the generated images.

\subsection{Quantitative Results}
\begin{figure}[ht]
    \centering
    \includegraphics[width=0.5\textwidth]{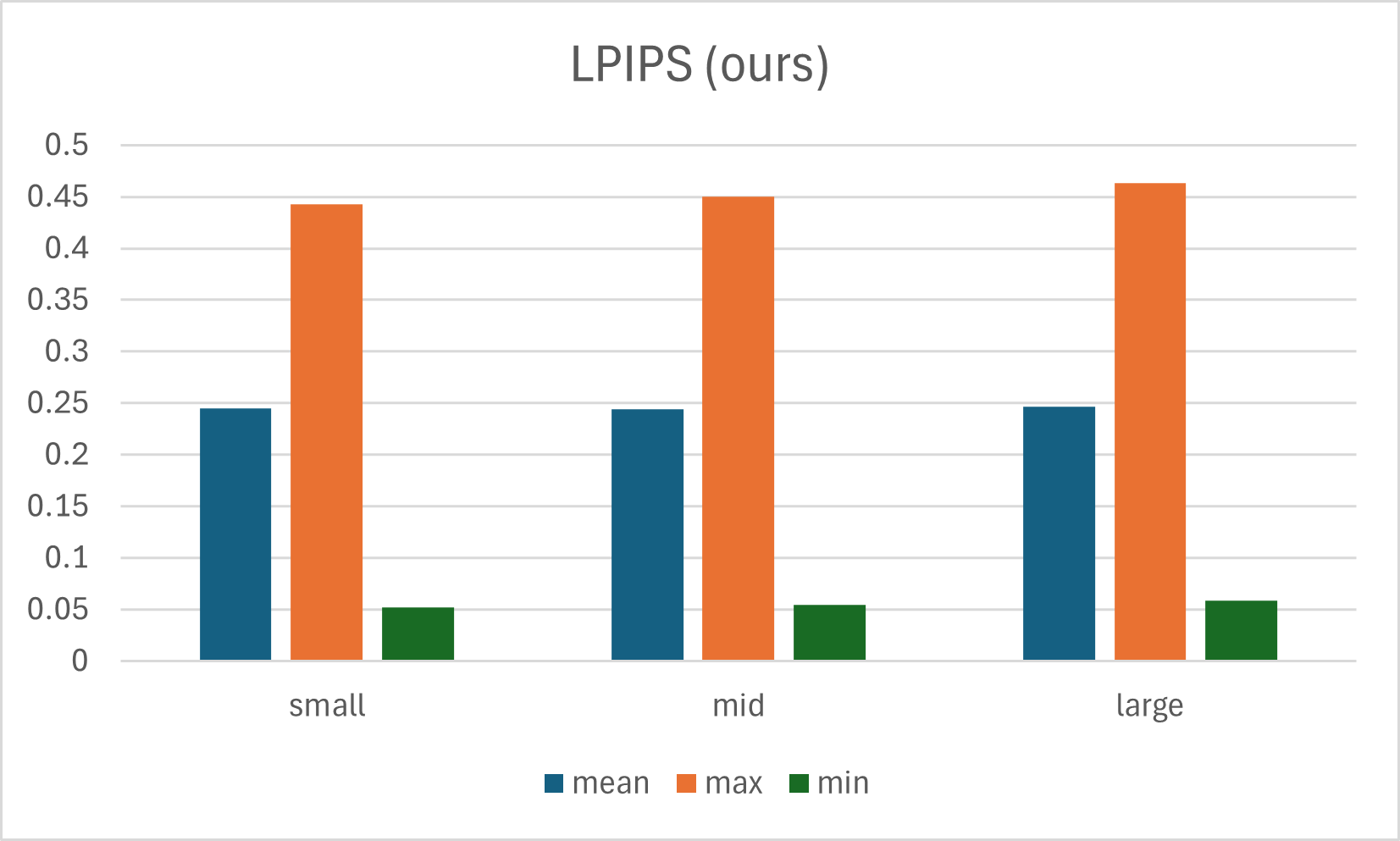}
    \caption{LPIPS scores for our model. The score closely resembles the baseline score. Ours does not match the baseline by a small margin.}
    \label{fig:lpip-ours}
\end{figure}

\begin{figure}[ht]
    \centering
    \begin{subfigure}[b]{0.5\textwidth}
        \centering
        \includegraphics[width=\linewidth]{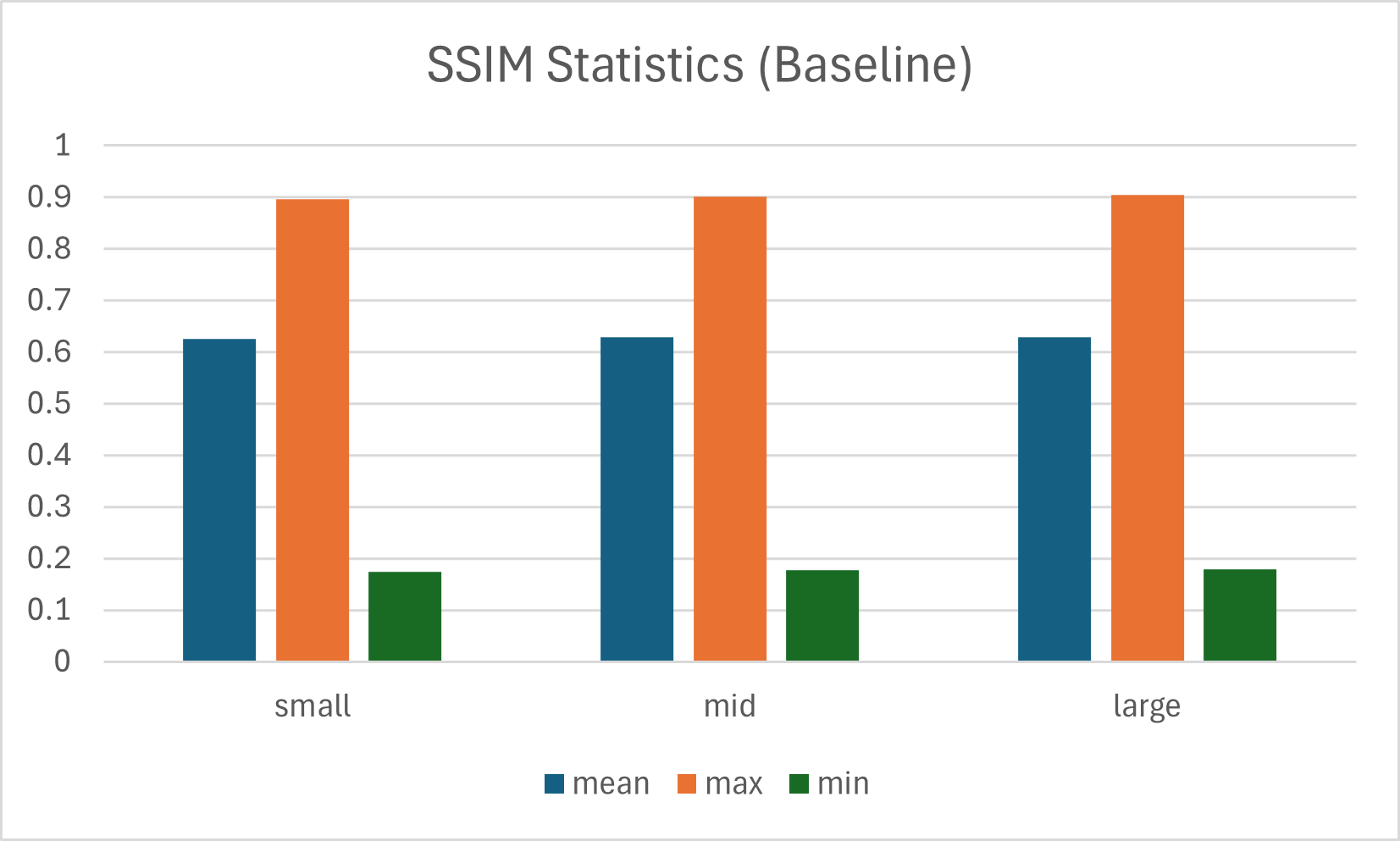}
    \end{subfigure}
    \begin{subfigure}[b]{0.5\textwidth}
        \centering
        \includegraphics[width=\linewidth]{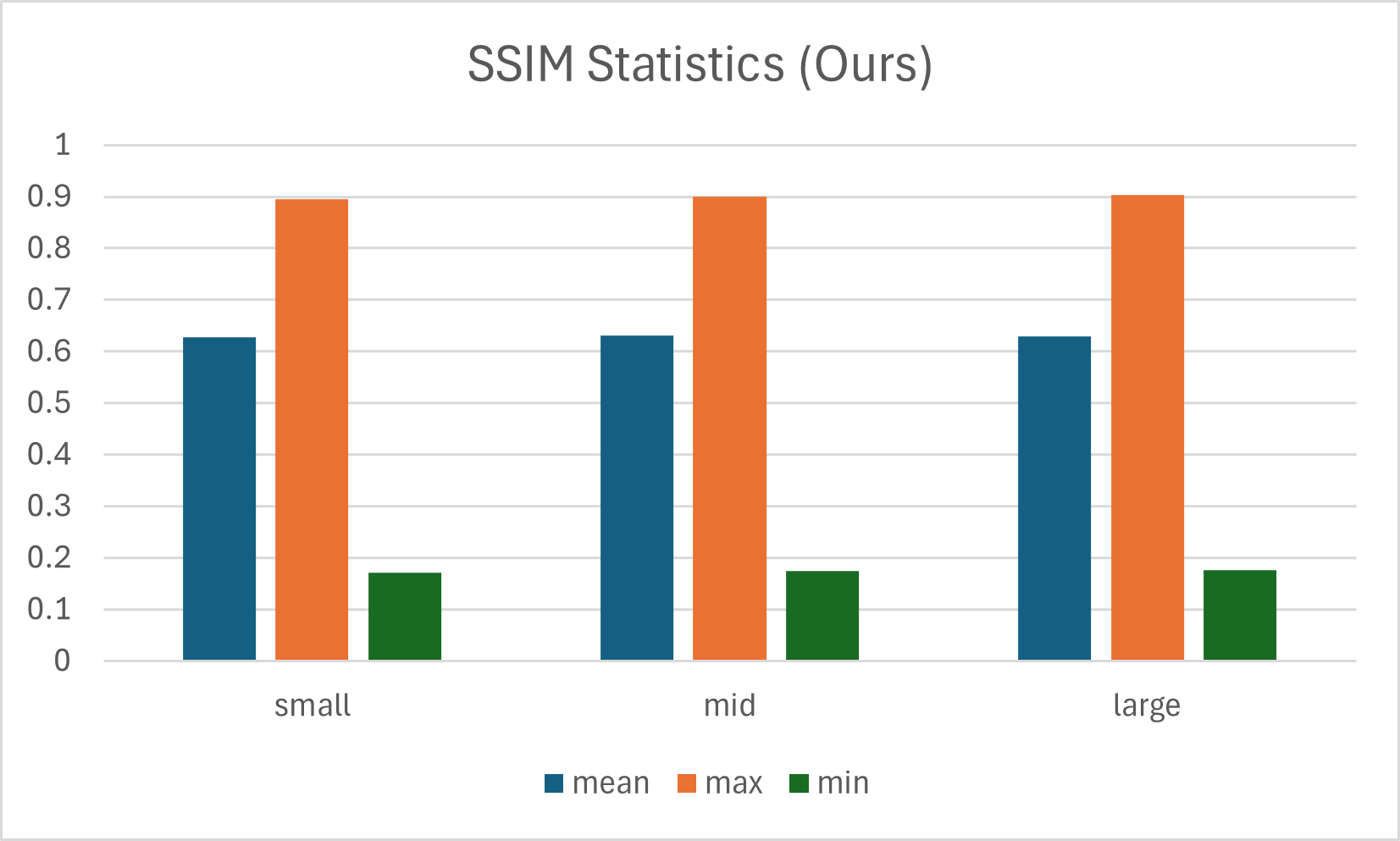}
    \end{subfigure}
    \caption{Side by side of the SSIM scores between the baseline (left) and ours (right). Ours had a high average score while the baseline maintained higher min/max scores.}
    \label{fig:ssim-scores}
\end{figure}

For quantitative evaluation, we employ SSIM \cite{wang2004image} and LPIPS \cite{zhang2018unreasonable} metrics in the paired setting. In the unpaired setting, realism is assessed using FID \cite{heusel2017gans} and KID \cite{binkowski2018demystifying} scores. Our implementation follows the evaluation paradigm \cite{morelli2023ladi}.
We evaluated 48 images, figure \ref{fig:ssim-scores} shows their resulting SSIM scores over different resolution sizes. Large was the training size. While the charts look very similar, ours improved the averaged SSIM score. The LPIPS scores were close between our results and the baseline see figure \ref{fig:lpip-ours}. 

\section{Future Work}
Glamtry can be improved by training its dependent models more extensively on our dataset. The reliance on clear head-to-wrist visibility, resulting from the use of OpenPose, causes our model to struggle with replacing watches in common close-up wrist photos. Training the model to accurately reproduce specular details is another avenue forward. This would improve the model for correct jewelry representation during try-on scenarios.



\subsection{acknowledgements}
We used Pranjal Datta's SSIM notebook \cite{ssim2023}

\section{Appendix}

{\small
\bibliographystyle{ieee}
\bibliography{egpaper_for_review}
}

\end{document}